\documentclass{article}
\usepackage{arxiv}
\usepackage{times}
\usepackage{url}
\usepackage{latexsym}
\usepackage{comment}
\usepackage{graphicx}
\usepackage{lipsum}
\usepackage{titlesec}
\usepackage{tcolorbox}
\usepackage{multirow}
\usepackage{caption}
\usepackage{subcaption}
\usepackage{amsmath}
\usepackage{xcolor,colortbl}
\usepackage[utf8]{inputenc} 
\DeclareCaptionType{Table}
\usepackage{tikz}
\usepackage{pifont}
\usepackage{amssymb}

\newcommand{\cmark}{\ding{51}}%

\title{\includegraphics[width=10cm]{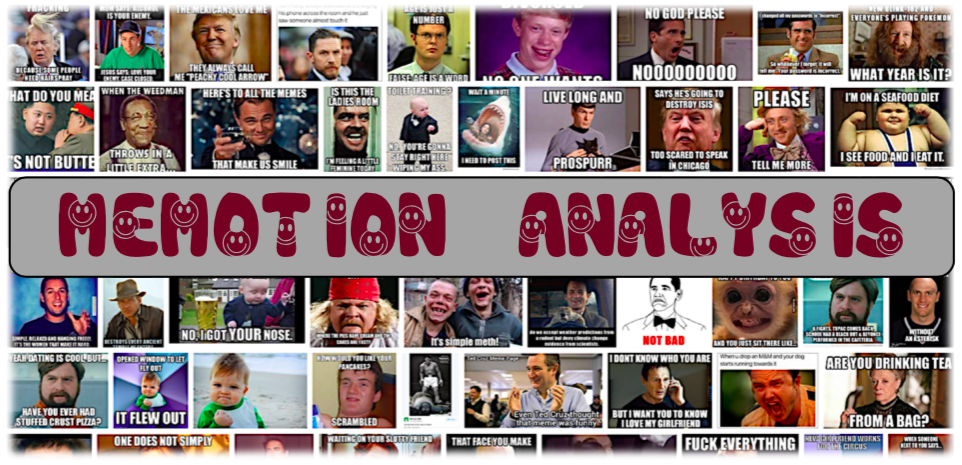}
\center
SemEval-2020 Task 8: Memotion Analysis- The Visuo-Lingual Metaphor!
}

\author{
Chhavi Sharma\textsuperscript{1} \quad 
Deepesh Bhageria\textsuperscript{1} \quad
William Scott\textsuperscript{2} \quad
Srinivas PYKL\textsuperscript{1} \quad 
{\bf Amitava Das\textsuperscript{3}}\quad\\
{\bf Tanmoy Chakraborty\textsuperscript{2}}\quad
{\bf Viswanath Pulabaigari\textsuperscript{1}}
{\bf Bj{\"o}rn Gamb{\"a}ck\textsuperscript{4}}\quad
\\
\textsuperscript{1}IIIT Sri City, India \quad
\textsuperscript{2}IIIT Delhi, India \quad
\textsuperscript{3}Wipro AI Labs, India\\
\textsuperscript{4}NTNU, Norway \\
{\tt \textsuperscript{1}\{chhavi.s, deepesh.b17,srinivas.p, viswanath.p\}@iiits.in}\\
{\tt \textsuperscript{2}\{william18026, tanmoy\}@iiitd.ac.in} \\
{\tt\textsuperscript{3}{amitava.das2@wipro.com}, 
\textsuperscript{4}{gamback@ntnu.no} 
}}

\begin{document}

\maketitle
\begin{abstract}
Information on social media comprises of various modalities such as textual, visual and audio. NLP and Computer Vision communities often leverage only one prominent modality in isolation to study social media. However, computational processing of Internet memes needs a hybrid approach. The growing ubiquity of Internet memes on social media platforms such as Facebook, Instagram, and Twitter further suggests that we can not ignore such multimodal content anymore. To the best of our knowledge, there is not much attention towards meme emotion analysis. The objective of this proposal is to bring the attention of the research community towards the automatic processing of Internet memes. The task Memotion analysis released approx 10K annotated memes- with human annotated labels namely \textit{sentiment}(positive, negative, neutral), type of \textit{emotion}(sarcastic,funny,offensive, motivation) and their corresponding intensity.
The challenge consisted of three subtasks: 
sentiment (\textit{positive, negative, and neutral\/}) analysis of memes, 
overall emotion (\textit{humor, sarcasm, offensive, and motivational\/}) classification of memes, 
and classifying intensity of meme emotion.
The best performances achieved were F$_1$ (macro average) scores of 0.35, 0.51 and 0.32, respectively for each of the three subtasks. 
\end{abstract}
\section{Introduction }
\label{sec:introduction}
In the last few years, the growing ubiquity of Internet memes on social media platforms such as Facebook, Instagram, and Twitter has become a topic of immense interest. Memes are one of the most typed English words Sonnad~\cite{meme_swift} in recent times which are often derived from our prior social and cultural experiences such as TV series or a popular cartoon character (think: “One Does Not Simply” - a now immensely popular meme taken from the movie \textit{Lord of the Rings}). These digital constructs are so deeply ingrained in our Internet culture that to understand the opinion of a community, we need to understand the type of memes it shares.~\cite{gal2016gets} aptly describes them as \textit{performative acts}, which involve a conscious decision to either support or reject an ongoing social discourse. \par
 The prevalence of hate speech in online social media is a nightmare and a great societal responsiblity for many social media companies. However, the latest entrant “Internet memes” ~\cite{williams2016racial} has doubled the challenge. When malicious users upload something offensive to torment or disturb people, it traditionally has to be seen and flagged by at least one human, either a user or a paid worker.
Even today, companies like Facebook and Twitter rely extensively on outside human contractors from different companies.But with the growing volume of multimodal social media it is becoming impossible to scale. The detection of offensive content on online social media is an ongoing struggle. OffenseEval ~\cite{ZampieriEA19} is a shared task which is being organized since the last two years at SemEval. But, detecting an offensive meme is more complex than detecting an offensive text -- as it involves visual cues and language understanding. This is one of the motivating aspects which encouraged us to propose this task.\par
  Analogous to textual content on social media, memes also need to be analysed and processed to extract the conveyed message. A few researchers have tried to automate the meme generation ~\cite{peirson2018dank,oliveira2016one} process, while a few others tried to extract its inherent sentiment ~\cite{french2017image} in the recent past. Nevertheless, a lot more needs to be done to distinguish their finer aspects such as type of humor or offense.\par
  The paper is organised as follows: The proposed task is described in Section~\ref{sec:tasks}. Data collection and data distribution is explained in Section~\ref{sec:Data collection} while Section~\ref{sec:Baseline_model} demonstrates the baseline model. Section~\ref{sec:Evaluation Metric} shows the reason for considering Macro F1 as evaluation metric. In Section~\ref{sec:system Analysis}, participants and the top performing models are discussed in detail. Section~\ref{sec:result} shows the results, analysis and the takeaway points from Memotion 1.0. Related work is described in Section \ref{sec:Related work}. Finally,  we  summarise  our  work  by highlighting the insights derived along-with the further scope and open ended pointers in section~\ref{sec:conclusion}.
 
\begin{figure*}[!ht]
\begin{minipage}[!t]{0.3\linewidth}
\centering
\includegraphics[scale=0.6]{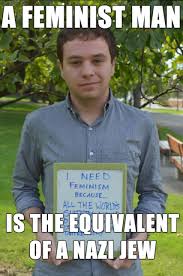}
\caption{A sarcastic and humorous meme on a feminist man. Here, the text is enough to get humor punchline.}
\label{fig:feminist}
\end{minipage}
\hspace{0.4cm}
\begin{minipage}[!t]{0.3\linewidth}
\centering
\includegraphics[scale=0.5]{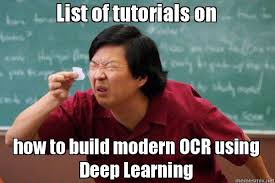}
\caption{A sarcastic meme on unavailability of deep learning based OCR materials on internet. The extreme shortage of tutorials is conveyed by the man in the meme through the imagery of attempting to read a small piece of paper.}
\label{fig:ocr}
\end{minipage}
\hspace{0.4cm}
\begin{minipage}[!t]{0.3\linewidth}
\centering
\includegraphics[scale=0.15]{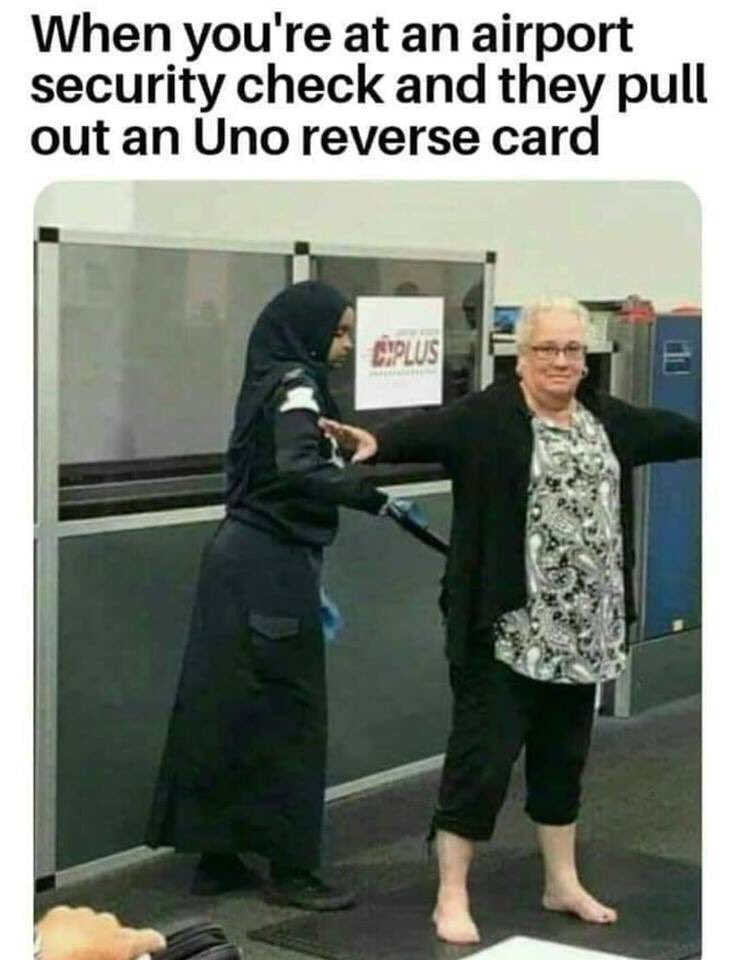}
\caption{An offensive meme on woman dressed in Hijab. It is difficult to label this as offensive until one makes the correlation between the biased opinion towards a particular religion and role reversal for the act of frisking. }
\label{fig:offensive}
\end{minipage}
\end{figure*}
\vspace{-7mm}

\section{The Memotion Analysis Task} \label{sec:tasks}

Memes typically induce humor and strive to be relatable. Many of them aim to express solidarity during certain life phases and thus, to connect with their audience. Some memes are directly humorous whereas others go for sarcastic dig at daily life events. Inspired by the various humorous effects of memes, we propose three task as follows: 
\begin{itemize}
\item Task \textbf{A}- \textbf{Sentiment Classification}: Given an Internet meme, the first task is to classify it as positive, negative or neutral meme.
\item Task \textbf{B}- \textbf{Humor Classification}: Given an Internet meme, the system has to identify the type of emotion expressed. The categories are \textit{sarcastic}, \textit{humorous}, \textit{motivation} and \textit{offensive} meme. A meme can have more than one category. For instance, Fig.\ref{fig:offensive} is an offensive meme but sarcastic too.  
\item Task \textbf{C}- \textbf{Scales of Semantic Classes}: The third task is to quantify the extent to which a particular effect is being expressed. Details of such quantifications is reported in the Table \ref{tab:LikertScale}. 
\end{itemize}
We have released 10K human annotated Internet memes labelled with semantic dimensions namely \textit{sentiment}, and type of humor that is, \textit{sarcastic}, \textit{humorous}, or \textit{offensive} and \textit{motivation} with their corresponding intensity. The humor types are further quantified on a Likert scale as in Table~\ref{tab:LikertScale}. The data-set will also contain the extracted captions/texts from the memes. 

\begin{table}[t!]
    \centering
    \resizebox{0.7\columnwidth}{!}{%
    \begin{tabular}{ccccc}
    \hline
        & \textbf{sarcastic} &\textbf{ humorous} & \textbf{offensive} &  \textbf{Motivation}  \\ \hline
         \textit{not} (0) & \includegraphics[scale=0.13]{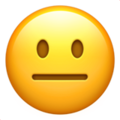} & \includegraphics[scale=0.13]{neutral-face_1f610.png}& \includegraphics[scale=0.13]{neutral-face_1f610.png} &
         \includegraphics[scale=0.13]{neutral-face_1f610.png}\\
        
         \textit{slightly} (1) & \includegraphics[scale=0.13]{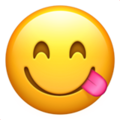} & \includegraphics[scale=0.13]{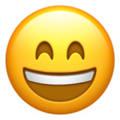}   & \includegraphics[scale=0.13]{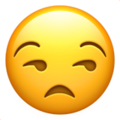} &
         \includegraphics[scale=0.18]{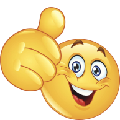}\\
        
         \textit{mildly} (2)& \includegraphics[scale=0.13]{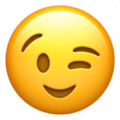} & \includegraphics[scale=0.13]{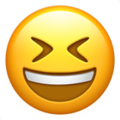} & \includegraphics[scale=0.13]{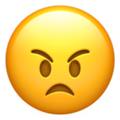} &
         \textit{NA}\\
         
         \textit{very} (3)& \includegraphics[scale=0.13]{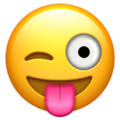}& \includegraphics[scale=0.13]{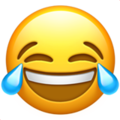}& \includegraphics[scale=0.13]{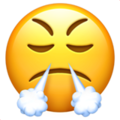} &
         \textit{NA} \\ \hline
    \end{tabular}
    }
    
    \captionof{Table}{Semantic classes for the Memotion Analysis}
    \label{tab:LikertScale}

\end{table}

\subsection{Memotion Analysis - Breaking the Ice!}
Taking into account the sheer volume of photos shared each day on Facebook, Twitter, and Instagram, the number of languages supported on our global platform, and the variations of the text, the problem of understanding text in images is quite different from those solved by traditional optical character recognition (OCR) systems, which recognize the characters but don't understand the context of the associated image~\cite{Rosetta}. 

For instance, the caption for the Fig.\ref{fig:feminist} is sufficient to sense the sarcasm or even dislike towards a feminist man. The image has no significant role to play in this case and the provided text is good enough to sense the pun. But, in Fig.\ref{fig:ocr}, the final punchline on the unavailability of Deep Learning based OCR tutorials is dependent on the man's expression in the meme who is trying to read a small piece of paper - the facial expression of the man aids in interpreting the struggle to find OCR tutorials online. To derive the intended meaning, someone needs to establish an association between the provided image and the text. If we solely process the caption, we will lose the humor and also the intended meaning (lack of tutorials). In Fig.\ref{fig:offensive}, to establish the sense of provided caption along with the racism against the middle east woman, we need to process the image, caption and dominant societal beliefs.
\section{Dataset}
\label{sec:Data collection}
To understand the complexity of memes, as discussed in the prior sections, it is essential to collect memes from different categories, with varying emotion classes. Details of preparing the data-set is presented below: 

\begin{figure*}[!tbh]
\centering
\includegraphics[width=1\textwidth]{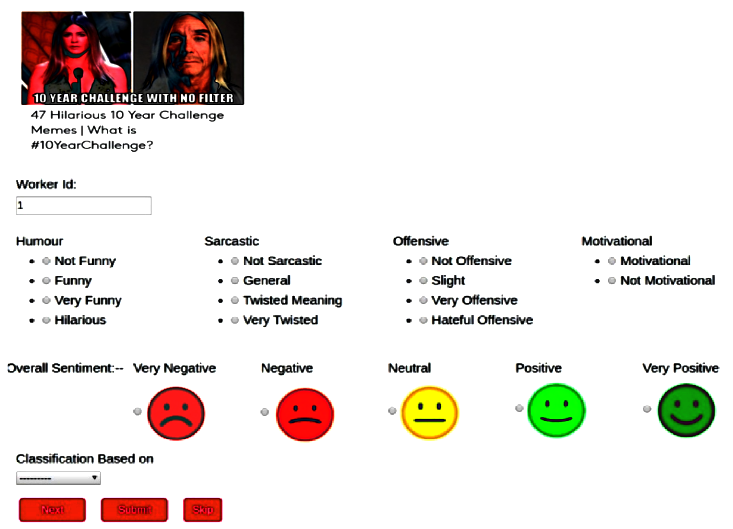}
\caption{Depiction of web interface, for crowd-sourced annotations: Annotation template page, with options provided for associated emotions and  their intensity levels, overall sentiment and the data-modality option, the annotation is based on.}
\label{fig:page 2}
\end{figure*}
\begin{itemize}
    \item \textbf{Data collection:} We identified a total of 52 unique and globally popular categories, for example, Hillary, Trump, Minions, Baby godfather, etc., for downloading the meme data. The meme (images) were downloaded using Google images search service, with the help of a browser extension tool called as \textit{fatkun batch downloader} \footnote{https://chrome.google.com/webstore/detail/fatkun-batch-download-ima/nnjjahlikiabnchcpehcpkdeckfgnohf?hl=en}. It provided a simple yet effective means to scrape a large number of memes, relevant for our purpose. To avoid any copyright issue in this, we have collected memes which are available in public domain along with their URLs, and added that information as additional meta-data in our data-set as well.
    \item \textbf{Filtering:} The memes are filtered keeping the following constraints into perspective:
    \begin{itemize}
    \item The meme must contain clear background picture, along-with an embedded textual content.
    \item Memes with only English language text content are considered for this study. 
    \end{itemize}

    \item\textbf{ Annotation:} For getting our data-set of 14k samples annotated, we reached out to Amazon Mechanical Turk (AMT) workers, to annotate the emotion class labels as \textit{Humorous, Sarcasm, Offensive, Motivation}  and quantify the intensity to which a particular effect of a  class is expressed, along-with the overall sentiments (very negative, negative, neutral, positive, very positive).
    \end{itemize}

Emotion about memes highly depends upon an individual's perception of an array of aspects within society, and could thus vary from one person to another. This phenomenon is called as "Subjective Perception Problem" as noted in ~\cite{subjective}. To address this challenge, the annotation process is performed multiple times ie. each sample is provided to 5 annotators, and the final annotations are adjudicated based upon majority voting scheme. The filtered and the annotated data-set comprises to the size of 9871 data samples. 
In addition to these aspects, textual content plays a pivotal role in ascertaining the emotion of a meme.


To understand the textual content from the memes, text has been extracted using Google vision OCR APIs. The extracted text was not completely accurate, therefore AMT workers were asked to provide the rectified text against the given OCR extracted text, for the inaccurately extracted OCR text.\par
\begin{figure*}[t]
\centering
\includegraphics[width=1.0\textwidth]{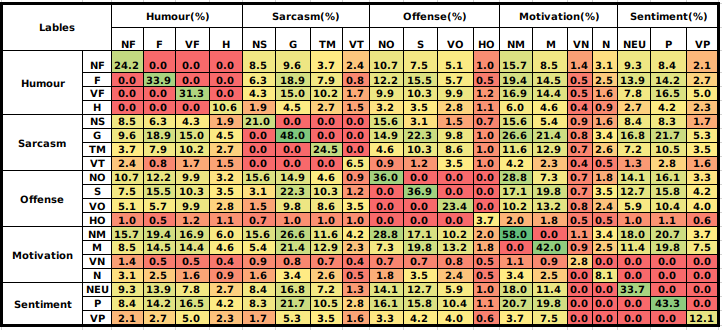}
\captionof{Table}{\textbf{Distribution of the overall data-set:}for categories \textit{Humour, Sarcasm, Motivation, Offense and Overall sentiment}, along-with their sub-categories (Abbreviations: NF: Not funny, F: Funny, VF: Very Funny, H: Humour; NS:Not Sarcastic, G: General sarcastic, TM: Twisted Meaning, VT: Very Twisted; NM: Not Motivational, M: Motivational; NO: Not Offensive, S: Slightly offensive, VO: Very Offensive, HO: Highly Offensive; VN: Very Negative, N: Negative, Neu: Neutral, P: Positive, VP: Very Positive)}
\label{fig:bigdata}
\end{figure*}

\begin{figure*}[t]
\centering
\includegraphics[width=\textwidth]{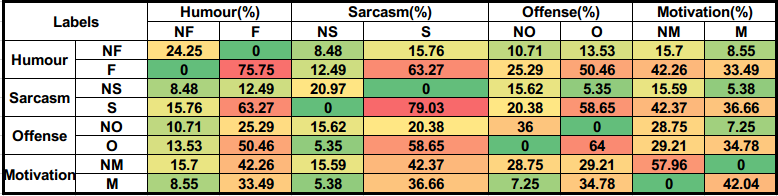}
\captionof{Table}{\textbf{Distribution of the data-set w.r.t Task 2}: for categories \textit{Humour, Sarcasm, Motivation, Offense and Overall sentiment} clubbed for two sub-categories at lower level of granularity. (NX-Abbreviation implying a 'Not' for a particular category X of emotions)}
\label{fig:smalldata}
\end{figure*}

\begin{figure*}[t]
\centering
\includegraphics[width=0.7\linewidth]{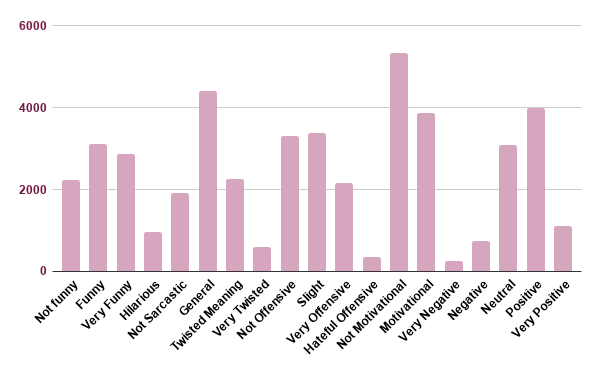}
\vspace{-3mm}
\caption{Plot depicting category-wise data distribution of meme emotion data-set [For eg. There are approx. 2200 memes in the data-set tagged as “Not funny”].}
\label{fig:datadistribution}
\end{figure*}

\textbf{Data Distribution:} The statistical summaries are provided in Table \ref{fig:bigdata} and \ref{fig:smalldata}. It is clear from the distribution of the data that there are significant overlapping emotions for the memes, which essentially validates the challenges discussed at the beginning. It can also be observed that majority of the memes are sarcastic in nature. Interestingly, most of the funny memes fall under the category of sarcastic class. Simultaneously, another noteworthy observation is that a significant number of memes are both motivational and offensive.\par
For the challenge, 1K samples were provided as trial data, 6992 samples as training data while 1879 samples as test data.

\section{Baseline Model}
\label{sec:Baseline_model}
Memes are the types of multi-modal content, wherein the graphical content and textual messages are self-sufficient to convey some meaning. But within the context of their dissemination, it is a specially tailored idea, that is designed to be propagated. Such complex ideas cannot be conveyed as effectively by any of the constituent data modality, as by their combination. In order to fully address the system modeling tasks that use such data, it is imperative to study the efficacy of individual content modality ie. image or text as well as their combination.
\subsection{Using only the textual features}
\label{sec:text}
\begin{figure*}[t]
\centering
\includegraphics[width=\linewidth]{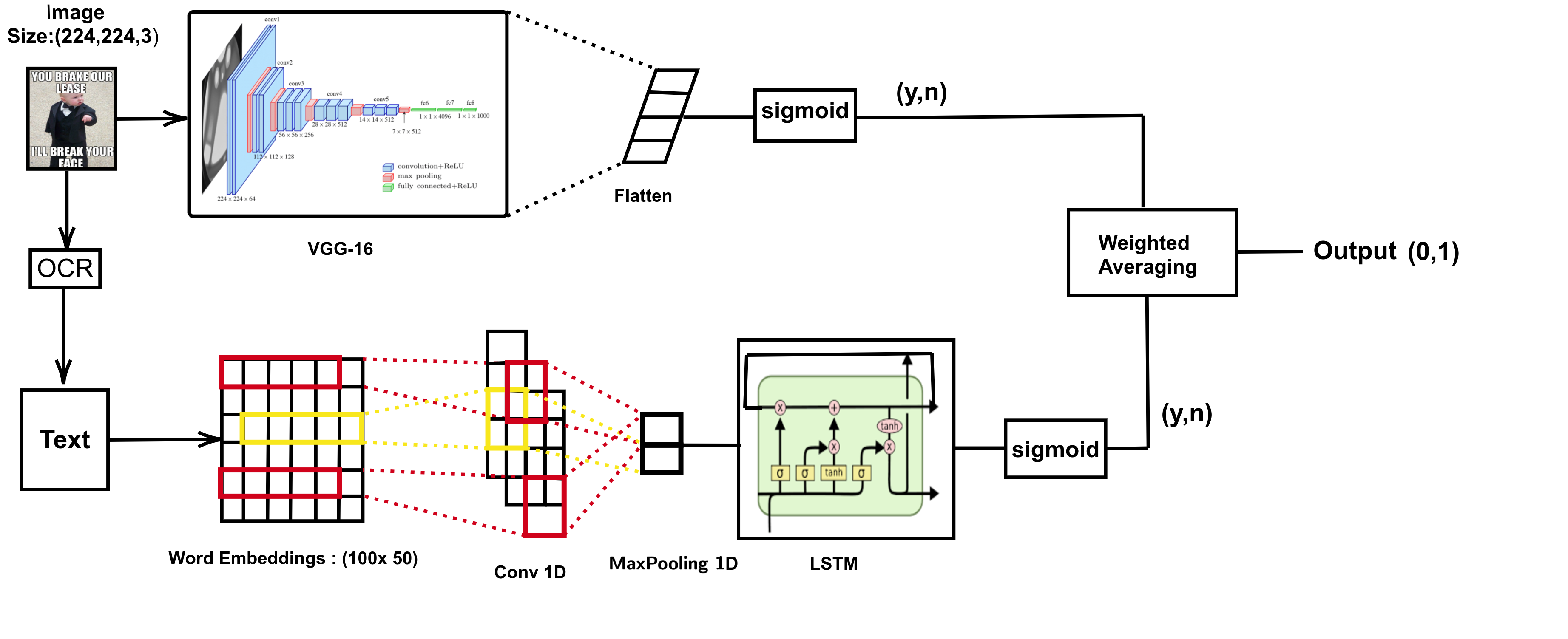}
\caption{Depiction of a multi-label classification system, employed towards emotion class classification. The system performs weighted averaging of image-text outputs to evaluate the class-wise scores.}
\label{fig:sec33}
\end{figure*}

A meme can be expressed using varying textual contents, so as to convey different emotions. In some cases, different memes can have same images, but due to the different textual messages embedded in each of them, different sentimental reactions can be induced from all.  Recognition of the emotion induced in such memes would require accurate modelling of the textual influence. To evaluate automated emotion recognition from the meme textual content, we built text binary classifier as shown in the bottom half of Fig.\ref{fig:sec33}, to understand different classes of emotion. We have used 100-D pre-trained Glove word embeddings ~~\cite{glove} to generate word embeddings from text $emd(txt)$. These embeddings are given as input $x_i$ to the CNN, having 64 filters of size 1$\times$5 with Relu as activation function to extract the textual features. To reduce the dimension of number of parameters generated by CNN layer we have used 1D maxpooling layer of size 2.  Weighted CNN output is given as input to LSTM where we get a feature vector  $s_{t}$. $s_{t}$ is fed to fully-connected layer, and activation function sigmoid is used to classify the text with binary cross-entropy as a loss function.





 \begin{figure*}[t]
\centering
\includegraphics[width=\linewidth]{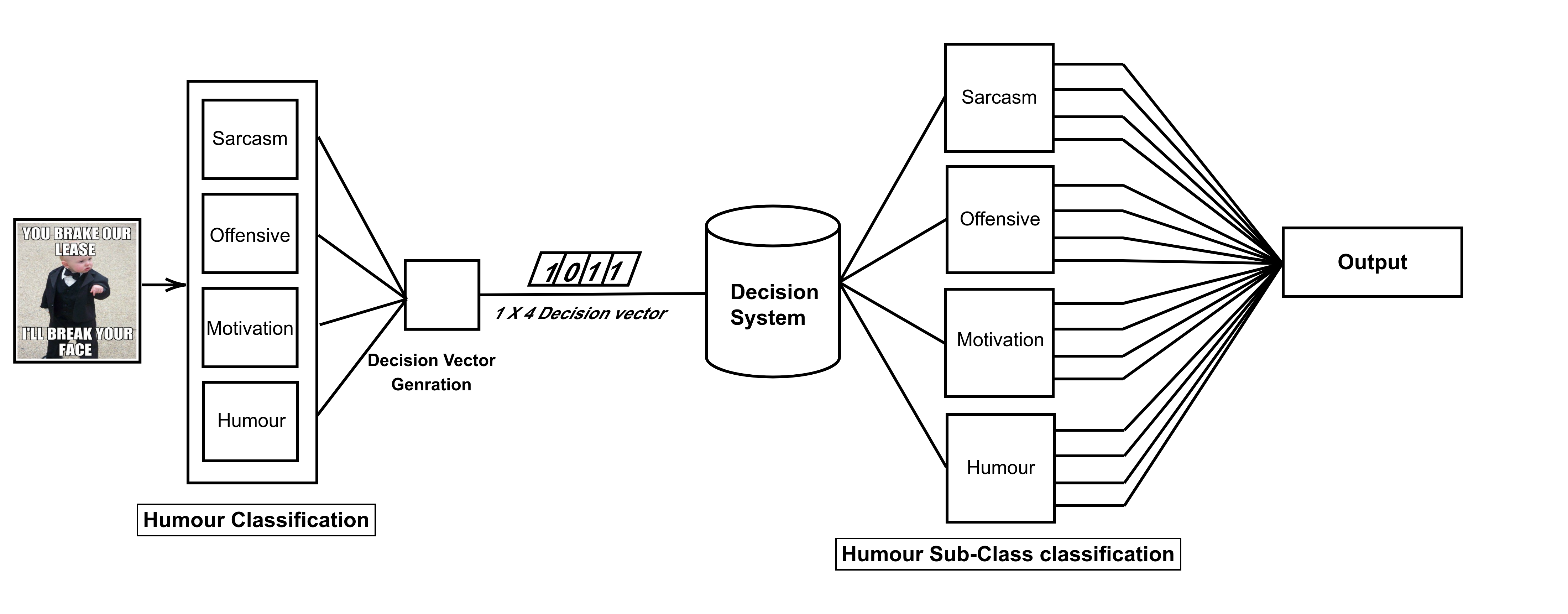}
\caption{A Multi-level system for the task of emotion intensity prediction (1$\times$14 dimensional), using the emotion class multi-label output (1$\times$4 dimensional). }
\label{fig:sec35}
\end{figure*}

\subsection{Visual cues for Memotion anslysis}
\label{sec:image}
To comprehend the significance of image in deducing the humour of a meme, we have used many pre-trained models like VGG-16~\cite{Simonyan2015VeryDC}, ResNet-50~\cite{DBLP:journals/corr/HeZRS15}, AlexNet~\cite{NIPS2012_4824}to extract the features of an image ,but  VGG-16 has given better features in comparison to other networks  due to it's capability of extracting both high and low level features. To maintain uniformity in images, we have resized the image into 224$\times$224$\times$3 from the original Image $I$, the resized image $X_{i}$ is fed into vgg16 as input for feature extraction. Extracted feature $Y_{i}$ from vgg16 for the given meme is flattened by flatten layer. Flatten output $Y_{i'}$ is fed to a fully connected network, and sigmoid function is used for classification with a loss function of binary cross entropy. 




\subsection{Visuo-Lingual modeling for Memotion}
\label{sec:text+image_classifier}

The information contained within a meme in any mode whether text or image, needs to be analyzed for recognizing the emotion associated with it. Analyses of the results obtained show that features extracted from Section ~\ref{sec:image} or~\ref{sec:text} alone are not sufficient to recognize the emotion of a meme. So we created the classifier for leveraging the combination of both image and text based model training, by performing \textit{weighted averaging} as shown in Fig.\ref{fig:sec33}, which resulted in better predictions. The model, shown in Fig.\ref{fig:sec33}, predicts the output for each class as $I_{1}$ \& $T_{1}$ for sarcastic, $I_{2}$ \& $T_{2}$ for humour, $I_{3}$ \& $T_{3}$ for offense and $I_{4}$ \& $T_{4}$ for motivation emotions, where $I_{i}$ is for image based model and $T_{i}$ is for text based model. To combine predicted probabilities of image and text, we have used softmax function $X_i$ with a weighted average of the obtained probabilities of image and text classifier. In our work we have used weighted average for scaling the predicted output and find the threshold $H$ to generate final 1$\times$4 output vector to show how a meme is classified into multiple classes, where $H$ is the average of image and text based network outputs, across the data-set size. The final output  $O$, which generates a 1$\times$4 vector for a given meme, is used as \textit{decision vector}  to activate the next level of emotion subclass to understand the intensity of emotion with respect to parent class.

\begin{table}[hbt!]
\centering
\resizebox{0.5\textwidth}{!}{%
\begin{tabular}{|c|c|c|c|c|}
\hline
\multirow{2}{*}{\textcolor{blue}{\textbf{Task B}}} & \multirow{2}{*}{\textcolor{blue}{\textbf{Classes}}} & \multicolumn{3}{c|}{\cellcolor{green}\textbf{Macro F1 score}}         \\ \cline{3-5} 
                                &                                   & \cellcolor{pink!80}\textbf{Image} & \cellcolor{pink!80}\textbf{Text} & \cellcolor{pink!80}\textbf{Image+Text} \\ \hline
{\textcolor{red}{\textbf{Task A}}}                         & \cellcolor{green!40}\textbf{Sentiment Analysis}       &  \cellcolor{green!40}\textbf{0.18          } & \cellcolor{green!40} \textbf{0.20 }         & \cellcolor{green!40} \textbf{0.21}                \\ \hline
\multirow{5}{*}{\textcolor{red}{\textbf{Task B}}}         & Humour                            & 0.48           & 0.49          & 0.51                \\ \cline{2-5} 
                                & Sarcasm                           & 0.51           & 0.53          & 0.50                \\ \cline{2-5} 
                                & Offense                           & 0.43           & 0.42          & 0.49                \\ \cline{2-5} 
                                & Motivation                        & 0.42           & 0.47          & 0.49                \\ \cline{2-5} 
                                & \cellcolor{green!40}\textbf{Average Score}            & \cellcolor{green!40} \textbf{0.46  }         & \cellcolor{green!40} \textbf{0.48}          & \cellcolor{green!40} \textbf{0.50 }               \\ \hline
\multirow{5}{*}{\textcolor{red}{\textbf{Task C}}}         & Humour                            & 0.21           & 0.23          & 0.24                \\ \cline{2-5} 
                                & Sarcasm                           & 0.24           & 0.25          & 0.24                \\ \cline{2-5} 
                                & Offense                           & 0.16           & 0.19          & 0.23                \\ \cline{2-5} 
                                & Motivation                        & 0.36           & 0.42          & 0.48                \\ \cline{2-5} 
                                & \cellcolor{green!40}\textbf{Average Score}            & \cellcolor{green!40} \textbf{0.24 }          & \cellcolor{green!40} \textbf{0.27   }       & \cellcolor{green!40} \textbf{0.30  }              \\ \hline
\end{tabular}
}
\captionof{Table}{Macro F1 score comparison for task-wise and class-wise (for task B and C) and their averages. The results are reported for evaluations for inputs as: image, text and the combination of both (image+text).}
\label{tab:result}
\end{table}

\subsection{Predicting the Intensity of memes}
To understand the intensity of an individual emotion associated with a meme,  we have created a multilabel classifier with two levels where at the first level meme is classified as sarcastic, offensive, motivational and humorous and the second level predicts the intensity of a particular class depending upon the vector generated at the first level, as depicted in Fig.\ref{fig:sec35}. Obtain predicted output in a vector form, of size 1$\times$4 from first level, i.e., emotion classification. In Fig.\ref{fig:sec35}, the decision system takes decision vector and activates the next multi-class classifier depending upon the value of the corresponding class in the vector. As we have a total of 14 classes (4 each of humour, sarcasm and offense while 2 of motivation), the final output that is the predicted intensity of each class will be a vector of size 1$\times$14.\par
The performance of the system is shown in Table \ref{tab:result} considering the image, text and the combination of both image and text.
\section{Evaluation Metric}
\label{sec:Evaluation Metric}
The challenge comprises of classifying the sentiment and emotion associated with a meme. The task A is a multi-class problem involved in identifying the sentiment (positive, negative, neutral) associated with a meme while the other 2 tasks B and C are multi-label classification problem associated with emotion detection. There are various evaluation metrics for multi-class and multi-label classification problem such as hamming loss, exact match ratio, macro/micro F1 score etc. The most used metric for this kind of problem is hamming loss which is evaluated as the fraction of wrongly classified label to the total number of labels. As our problem deals with different emotions associated with a meme, we have used macro F1 score that will help us to evaluate and analyse the individual class performance. 
\section{Participation and top performing systems}
\label{sec:system Analysis}
The challenge was a great success, involving total of 583 participants, with varying submissions in different tasks comprising of 31, 26 and 23 submissions in Task A, Task B and Task C respectively where in evaluation phase, a user is allowed for 5 submissions per day. 27 teams submitted the system description paper. A brief description of the task wise top performing models is shown below.
\subsection{Top 3 Task A systems @Memotion}
\begin{itemize}
\item \textbf{IITK Vkeswani:} Employed wide variety of methods, ranging from a simple linear classifier such as FFNN, Naive Bayes to transformers like MMBT~\cite{rahman2019mbert} and BERT~\cite{devlin-etal-2019-bert}. Implemented the model considering only text and the combination of image and text.
   
\item \textbf{Guoym:} Used ensembling Method considering the textual features extracted using Bi-GRU, BERT~\cite{devlin-etal-2019-bert}, or ELMo~\cite{peters2018deep}, image features
extracted by Resnet50~\cite{DBLP:journals/corr/HeZRS15} network and fusion features of text and images.

\item \textbf{Aihaihara:} Implemented the model that is a concatenation of visual and textual features obtained from n-gram language model and VGG-16~\cite{Simonyan2015VeryDC} pretrained model respectively.
\end{itemize}
\subsection{Top 3 Task B and Task C systems @Memotion}
\begin{itemize}
\item \textbf{UPB George:} In order to extract most salient features from text input, they opted to use the ALBERT~\cite{lan2019albert}model while VGG -16~\cite{Simonyan2015VeryDC} is used for  extracting the visual features from image input. To determine the humour associated with a meme, they have concatenated the visual and textual features followed by an output layer of softmax. 
  
\item \textbf{Souvik Mishra Kraken:} Applied Transfer learning by using hybrid
neural Naïve-Bayes Support Vector Machine and logistic regression for solving the task of humour classification and significant score.

\item \textbf{Hitachi:} They have  proposed
simple but effective MODALITY ENSEMBLE that incorporates visual and textual deep-learning
models, which are independently trained, rather than providing a single multi-modal joint network. They fine-tuned four pre-trained visual models (i.e., Inception-ResNet~\cite{DBLP:journals/corr/SzegedyIV16},
Polynet~\cite{zhang2016polynet}, SENet~\cite{DBLP:journals/corr/abs-1709-01507}, and PNASNet~\cite{DBLP:journals/corr/abs-1712-00559}) and four textual models (i.e., BERT~\cite{devlin-etal-2019-bert}, GPT-2~\cite{noauthororeditor}, Transformer-XL~\cite{dai2019transformerxl},
and XLNet~\cite{DBLP:journals/corr/abs-1906-08237}), followed by the fusion of their predictions by ensemble methods to effectively capture cross-
modal correlations.
\end{itemize}


\begin{table}[!tbh]
\centering
\resizebox{0.8\textwidth}{!}{%
\begin{tabular}{|c|c|c|}
\hline
\rowcolor[HTML]{009901} 
\multicolumn{3}{|c|}{\cellcolor[HTML]{009901}{\color[HTML]{FFFFFF} \textbf{Task-A Sentiment Analsysis}}}                                                   \\ \hline
\rowcolor[HTML]{FFCE93} 
{\color[HTML]{656565} \textbf{Participant / Team}}            & {\color[HTML]{656565} \textbf{Macro-F1}} & {\color[HTML]{656565} \textbf{Comparison with baseline(+/-)}} \\ \hline
\textbf{Vkeswani IITK}                     & \textbf{0.35466}                         & {\color[HTML]{009901} \textbf{(+)0.13701}}                    \\ \hline
\textbf{Guoym}                       & \textbf{0.35197}                         & {\color[HTML]{009901} \textbf{(+)0.13432}}                    \\ \hline
\textbf{Aihaihara}                          & \textbf{0.35017}                         & {\color[HTML]{009901} \textbf{(+)0.13252}}                    \\ \hline
\textbf{Sourya Diptadas}                     & \textbf{0.34885}                         & {\color[HTML]{009901} \textbf{(+)0.13120}}                    \\ \hline
\textbf{Irina Bejan}                         & \textbf{0.34755}                         & {\color[HTML]{009901} \textbf{(+)0.12990}}                    \\ \hline
\textbf{Sabino INGEOTEC}                   & \textbf{0.34689}                         & {\color[HTML]{009901} \textbf{(+)0.12924}}                    \\ \hline
\textbf{U.Walinska Urszula}      & \textbf{0.34639}                         & {\color[HTML]{009901} \textbf{(+)0.12874}}                    \\ \hline
\textbf{Souvik Mishra Kraken}               & \textbf{0.34627}                         & {\color[HTML]{009901} \textbf{(+)0.12862}}                    \\ \hline
\textbf{Lb732 SESAM}                       & \textbf{0.34600}                         & {\color[HTML]{009901} \textbf{(+)0.12835}}                    \\ \hline
\textbf{Li Zhen hit-mitlab}               & \textbf{0.34583}                         & {\color[HTML]{009901} \textbf{(+)0.12818}}                    \\ \hline
\textbf{George.Vlad Eduardgzaharia UPB}   & \textbf{0.34539}                         & {\color[HTML]{009901} \textbf{(+)0.12774}}                    \\ \hline
\textbf{Xiaoyu}                     & \textbf{0.34522}                         & {\color[HTML]{009901} \textbf{(+)0.12757}}                    \\ \hline
\textbf{HonoMi Hitachi}                    & \textbf{0.34145}                         & {\color[HTML]{009901} \textbf{(+)0.12381}}                    \\ \hline
\textbf{Gundapu Sunil}                       & \textbf{0.33915}                         & {\color[HTML]{009901} \textbf{(+)0.12150}}                    \\ \hline
\textbf{Surya}                              & \textbf{0.33861}                         & {\color[HTML]{009901} \textbf{(+)0.12096}}                    \\ \hline
\textbf{Jy930 Rippleai}                    & \textbf{0.33732}                         & {\color[HTML]{009901} \textbf{(+)0.11967}}                    \\ \hline
\textbf{Prhlt upv}                          & \textbf{0.33555}                         & {\color[HTML]{009901} \textbf{(+)0.11790}}                    \\ \hline
\textbf{Lyr123 Mem3210}                    & \textbf{0.33388}                         & {\color[HTML]{009901} \textbf{(+)0.11623}}                    \\ \hline
\textbf{YuanLi95 YNU-HPCC}                 & \textbf{0.33266}                         & {\color[HTML]{009901} \textbf{(+)0.11501}}                    \\ \hline
\textbf{Sanath97 TDResearch}               & \textbf{0.33228}                         & {\color[HTML]{009901} \textbf{(+)0.11464}}                    \\ \hline
\textbf{NLPU iowa NLP@UIowa}              & \textbf{0.32873}                         & {\color[HTML]{009901} \textbf{(+)0.11109}}                    \\ \hline
\textbf{Cosec}                              & \textbf{0.32869}                         & {\color[HTML]{009901} \textbf{(+)0.11104}}                    \\ \hline
\textbf{Mayukh memebusters}                & \textbf{0.32540}                         & {\color[HTML]{009901} \textbf{(+)0.10775}}                    \\ \hline
\textbf{Steve050798 NIT-Agartala-NLP-Team} & \textbf{0.32480}                         & {\color[HTML]{009901} \textbf{(+)0.10715}}                    \\ \hline
\textbf{Nowshed CSECU KDE MA}            & \textbf{0.32301}                         & {\color[HTML]{009901} \textbf{(+)0.10536}}                    \\ \hline
\textbf{Hg}                                 & \textbf{0.32291}                         & {\color[HTML]{009901} \textbf{(+)0.10526}}                    \\ \hline
\textbf{LT3}                                & \textbf{0.32206}                         & {\color[HTML]{009901} \textbf{(+)0.10441}}                    \\ \hline
\textbf{Abaruah IIITG-ADBU}                & \textbf{0.30780}                         & {\color[HTML]{009901} \textbf{(+)0.09015}}                    \\ \hline
\textbf{Sravani IS}                        & \textbf{0.29651}                         & {\color[HTML]{009901} \textbf{(+)0.07886}}                    \\ \hline
\textbf{Taha}                               & \textbf{0.29649}                         & {\color[HTML]{009901} \textbf{(+)0.07884}}                    \\ \hline
\textbf{Hamadanayel NAYEL}                 & \textbf{0.28748}                         & {\color[HTML]{009901} \textbf{(+)0.06983}}                    \\ \hline
\textbf{Saradhix Fermi}                    & \textbf{0.24780}                         & {\color[HTML]{009901} \textbf{(+)0.03015}}                    \\ \hline
\rowcolor[HTML]{9698ED} 
\textbf{Baseline}                          & \textbf{0.21765}                         & \textbf{}                                                     \\ \hline
\end{tabular}
}
\captionof{Table}{Team-wise results (Macro-F1) and their comparison with the base-line performance, for Task A—Sentiment Analysis [Comparison color code: Green--ahead of the base-line; Red--behind the base-line].}
\label{tab:taska}
\end{table}

\begin{table}[!tbh]
\centering
\resizebox{1.0\textwidth}{!}{%
\begin{tabular}{|c|c|c|c|c|c|c|}
\hline
\rowcolor[HTML]{036400} 
\multicolumn{7}{|c|}{\cellcolor[HTML]{036400}{\color[HTML]{FFFFFF} \textbf{Task B: Humour Classification (Macro F1 Score)}}}                                                                                                                                                                                                              \\ \hline
\rowcolor[HTML]{FFCE93} 
{\color[HTML]{656565} \textbf{Participant / Team}}            & {\color[HTML]{656565} \textbf{Humour}} & {\color[HTML]{656565} \textbf{Sarcasm}} & {\color[HTML]{656565} \textbf{Offense}} & {\color[HTML]{656565} \textbf{Motivation}} & {\color[HTML]{656565} \textbf{Average Score}} & {\color[HTML]{656565} \textbf{Comparison with Baseline(+/-)}} \\ \hline
{\color[HTML]{cc3399}\textbf{George.Vlad Eduardgzaharia UPB}}   & \textbf{0.51587}                       & {\color[HTML]{cc3399}\textbf{0.51590}}                       & \textbf{0.52250}                        & \textbf{0.51909}                           & \textbf{0.51834}                              & {\color[HTML]{32CB00} \textbf{+0.01813}}                      \\ \hline
\textbf{Guoym}                       & \textbf{0.51493}                       & \textbf{0.51099}                        & \textbf{0.51196}                        & \textbf{0.52065}                           & \textbf{0.51463}                              & {\color[HTML]{32CB00} \textbf{+0.01442}}                      \\ \hline
\textbf{Souvik Mishra Kraken}               & \textbf{0.51450}                       & \textbf{0.50415}                        & \textbf{0.51230}                        & \textbf{0.50708}                           & \textbf{0.50951}                              & {\color[HTML]{32CB00} \textbf{+0.00930}}                      \\ \hline
\textbf{Prhlt upv}                          & \textbf{0.50956}                       & \textbf{0.51311}                        & \textbf{0.50556}                        & \textbf{0.50912}                           & \textbf{0.50934}                              & {\color[HTML]{32CB00} \textbf{+0.00913}}                      \\ \hline
{\color[HTML]{cc3399}\textbf{Mayukh Memebusters}}                &{\color[HTML]{cc3399}\textbf{0.52992}}                       & \textbf{0.48481}                        & {\color[HTML]{cc3399}\textbf{0.52907}}                       & \textbf{0.49069}                           & \textbf{0.50862}                              & {\color[HTML]{32CB00} \textbf{+0.00841}}                      \\ \hline
\textbf{NLPU iowa NLP@UIowa}              & \textbf{0.51210}                       & \textbf{0.50389}                        & \textbf{0.50427}                        & \textbf{0.50896}                           & \textbf{0.50730}                              & {\color[HTML]{32CB00} \textbf{+0.00709}}                      \\ \hline
{\color[HTML]{cc3399}\textbf{Saradhix Fermi}    }                & \textbf{0.50214}                       & \textbf{0.49340}                        & \textbf{0.49648}                        & {\color[HTML]{cc3399}\textbf{0.53411}}                           & \textbf{0.50653}                              & {\color[HTML]{32CB00} \textbf{+0.00632}}                      \\ \hline
\textbf{Jy930 Rippleai}                    & \textbf{0.50035}                       & \textbf{0.48352}                        & \textbf{0.51589}                        & \textbf{0.52033}                           & \textbf{0.50502}                              & {\color[HTML]{32CB00} \textbf{+0.00481}}                      \\ \hline
\textbf{Cosec}                              & \textbf{0.50983}                       & \textbf{0.49471}                        & \textbf{0.49459}                        & \textbf{0.50327}                           & \textbf{0.50060}                              & {\color[HTML]{32CB00} \textbf{+0.00039}}                      \\ \hline
\rowcolor[HTML]{9698ED} 
\textbf{Baseline}              & \textbf{0.51185}                       & \textbf{0.50635}                        & \textbf{0.49114}                        & \textbf{0.49148}                           & \textbf{0.50021}                              & \textbf{}                                                     \\ \hline

\textbf{Steve050798 NIT-Agartala-NLP-Team} & \textbf{0.49247}                       & \textbf{0.50190}                        & \textbf{0.50533}                        & \textbf{0.49799}                           & \textbf{0.49942}                              & {\color[HTML]{FE0000} \textbf{-0.00079}}                      \\ \hline
\textbf{Sourya Diptadas}                     & \textbf{0.51387}                       & \textbf{0.49544}                        & \textbf{0.48635}                        & \textbf{0.49432}                           & \textbf{0.49750}                              & {\color[HTML]{FE0000} \textbf{-0.00271}}                      \\ \hline
\textbf{Hg}                                 & \textbf{0.48583}                       & \textbf{0.50017}                        & \textbf{0.47254}                        & \textbf{0.52218}                           & \textbf{0.49518}                              & {\color[HTML]{FE0000} \textbf{-0.00503}}                      \\ \hline
\textbf{Surya}                              & \textbf{0.50156}                       & \textbf{0.49949}                        & \textbf{0.47850}                        & \textbf{0.49831}                           & \textbf{0.49446}                              & {\color[HTML]{FE0000} \textbf{-0.00575}}                      \\ \hline
\textbf{Gundapu Sunil}                       & \textbf{0.50156}                       & \textbf{0.49949}                        & \textbf{0.47850}                        & \textbf{0.49831}                           & \textbf{0.49446}                              & {\color[HTML]{FE0000} \textbf{-0.00575}}                      \\ \hline
\textbf{Sanath97 TDResearch}               & \textbf{0.51269}                       & \textbf{0.47938}                        & \textbf{0.49905}                        & \textbf{0.48564}                           & \textbf{0.49419}                              & {\color[HTML]{FE0000} \textbf{-0.00602}}                      \\ \hline
\textbf{Nowshed CSECU KDE MA}            & \textbf{0.49272}                       & \textbf{0.48705}                        & \textbf{0.50480}                        & \textbf{0.49053}                           & \textbf{0.49377}                              & {\color[HTML]{FE0000} \textbf{-0.00644}}                      \\ \hline
\textbf{Lyr123 mem3210}                    & \textbf{0.48745}                       & \textbf{0.48948}                        & \textbf{0.48724}                        & \textbf{0.50872}                           & \textbf{0.49322}                              & {\color[HTML]{FE0000} \textbf{-0.00699}}                      \\ \hline
\textbf{Lb732 SESAM}                       & \textbf{0.46738}                       & \textbf{0.49180}                        & \textbf{0.51032}                        & \textbf{0.49910}                           & \textbf{0.49215}                              & {\color[HTML]{FE0000} \textbf{-0.00806}}                      \\ \hline
\textbf{HonoMi Hitachi}                    & \textbf{0.52136}                       & \textbf{0.44064}                        & \textbf{0.49116}                        & \textbf{0.51167}                           & \textbf{0.49121}                              & {\color[HTML]{FE0000} \textbf{-0.00900}}                      \\ \hline
\textbf{Vkeswani IITK}                     & \textbf{0.47352}                       & \textbf{0.50855}                        & \textbf{0.49993}                        & \textbf{0.47379}                           & \textbf{0.48895}                              & {\color[HTML]{FE0000} \textbf{-0.01126}}                      \\ \hline
\textbf{Hamadanayel NAYEL}                 & \textbf{0.48016}                       & \textbf{0.45595}                        & \textbf{0.48549}                        & \textbf{0.49105}                           & \textbf{0.47816}                              & {\color[HTML]{FE0000} \textbf{-0.02205}}                      \\ \hline
\textbf{INGEOTEC-sabino}                    & \textbf{0.47801}                       & \textbf{0.49920}                        & \textbf{0.45023}                        & \textbf{0.48177}                           & \textbf{0.47730}                              & {\color[HTML]{FE0000} \textbf{-0.02291}}                      \\ \hline
\textbf{Sabino INGEOTEC}                   & \textbf{0.47801}                       & \textbf{0.49920}                        & \textbf{0.45023}                        & \textbf{0.48177}                           & \textbf{0.47730}                              & {\color[HTML]{FE0000} \textbf{-0.02291}}                      \\ \hline
\textbf{Abaruah IIITG-ADBU}                & \textbf{0.47891}                       & \textbf{0.44920}                        & \textbf{0.42256}                        & \textbf{0.50957}                           & \textbf{0.46506}                              & {\color[HTML]{FE0000} \textbf{-0.03515}}                      \\ \hline
\textbf{LT3}                                & \textbf{0.47310}                       & \textbf{0.45572}                        & \textbf{0.40666}                        & \textbf{0.51410}                           & \textbf{0.46240}                              & {\color[HTML]{FE0000} \textbf{-0.03781}}                      \\ \hline
\textbf{IrinaBejan}                         & \textbf{0.46105}                       & \textbf{0.45450}                        & \textbf{0.50485}                        & \textbf{0.38748}                           & \textbf{0.45197}                              & {\color[HTML]{FE0000} \textbf{-0.04824}}                      \\ \hline
\textbf{Taha IUST}                         & \textbf{0.45473}                       & \textbf{0.45085}                        & \textbf{0.44529}                        & \textbf{0.43149}                           & \textbf{0.44559}                              & {\color[HTML]{FE0000} \textbf{-0.05462}}                      \\ \hline
\textbf{Xiaoyu}                     & \textbf{0.43376}                       & \textbf{0.44663}                        & \textbf{0.39965}                        & \textbf{0.48848}                           & \textbf{0.44213}                              & {\color[HTML]{FE0000} \textbf{-0.05808}}                      \\ \hline
\textbf{YuanLi95 YNU-HPCC}                 & \textbf{0.45756}                       & \textbf{0.44249}                        & \textbf{0.40310}                        & \textbf{0.29775}                           & \textbf{0.40023}                              & {\color[HTML]{FE0000} \textbf{-0.09998}}                      \\ \hline
\end{tabular}
}
\captionof{Table}{Team/Class--wise results (Macro-F1) and their comparison with the base-line performance, for Task B - Humour Classification.  Table is arranged in descending order ie. top-most row shows the winner while the $2^{nd}$ row tells about the score of $2^{nd}$ ranker. The scores and the team names highlighted in pink color shows the class wise best result. [Comparison color code: Green--ahead of the base-line; Red--behind the base-line].}
\label{tab:taskb}
\end{table}

\begin{table}[!tbh]
\centering
\resizebox{1.0\textwidth}{!}{%
\begin{tabular}{|c|c|c|c|c|c|c|}
\hline
\rowcolor[HTML]{009901} 
\multicolumn{7}{|c|}{\cellcolor[HTML]{009901}{\color[HTML]{FFFFFF} \textbf{Task C: Semantic Classification (Macro F1 Score)}}}                                                                                                                                                                                                            \\ \hline
\rowcolor[HTML]{FFCE93} 
{\color[HTML]{656565} \textbf{Participant / Team}}            & {\color[HTML]{656565} \textbf{Humour}} & {\color[HTML]{656565} \textbf{Sarcasm}} & {\color[HTML]{656565} \textbf{Offense}} & {\color[HTML]{656565} \textbf{Motivation}} & {\color[HTML]{656565} \textbf{Average Score}} & {\color[HTML]{656565} \textbf{Comparison with Baseline(+/-)}} \\ \hline
{\color[HTML]{cc3399}\textbf{Guoym}}                      & {\color[HTML]{cc3399}\textbf{0.27069}}                      & \textbf{0.25028}                        & \textbf{0.25761}                        & \textbf{0.51126}                           & \textbf{0.32246}                              & {\color[HTML]{009901} \textbf{(+)0.02157}}                    \\ \hline
\textbf{HonoMi Hitachi}                    & \textbf{0.26401}                       & \textbf{0.25378}                        & \textbf{0.24078}                        & \textbf{0.51679}                           & \textbf{0.31884}                              & {\color[HTML]{009901} \textbf{(+)0.01795}}                    \\ \hline
\textbf{George.vlad Eduardgzaharia UPB}   & \textbf{0.24874}                       & \textbf{0.25392}                        & \textbf{0.24688}                        & \textbf{0.51909}                           & \textbf{0.31716}                              & {\color[HTML]{009901} \textbf{(+)0.01627}}                    \\ \hline
\textbf{Jy930 Rippleai}                    & \textbf{0.25115}                       & \textbf{0.23783}                        & \textbf{0.25617}                        & \textbf{0.52033}                           & \textbf{0.31637}                              & {\color[HTML]{009901} \textbf{(+)0.01548}}                    \\ \hline
{\color[HTML]{cc3399}\textbf{Vkeswani IITK}}                    & \textbf{0.26171}                       &{\color[HTML]{cc3399}\textbf{0.25889}}& \textbf{0.26377}                        & \textbf{0.47379}                           & \textbf{0.31454}                              & {\color[HTML]{009901} \textbf{(+)0.01365}}                    \\ \hline
\textbf{Prhlt-upv}                          & \textbf{0.25634}                       & \textbf{0.24382}                        & \textbf{0.24815}                        & \textbf{0.50912}                           & \textbf{0.31436}                              & {\color[HTML]{009901} \textbf{(+)0.01347}}                    \\ \hline
\textbf{Hamadanayel NAYEL}                 & \textbf{0.25958}                       & \textbf{0.24406}                        & \textbf{0.26061}                        & \textbf{0.49105}                           & \textbf{0.31382}                              & {\color[HTML]{009901} \textbf{(+)0.01293}}                    \\ \hline
{\color[HTML]{cc3399}\textbf{Mayukh Memebusters}}                & \textbf{0.26127}                       & \textbf{0.23655}                        & {\color[HTML]{cc3399}\textbf{0.26512}}                       & \textbf{0.49069}                           & \textbf{0.31341}                              & {\color[HTML]{009901} \textbf{(+)0.01252}}                    \\ \hline
\textbf{Sourya Diptadas}                     & \textbf{0.26499}                       & \textbf{0.24498}                        & \textbf{0.24579}                        & \textbf{0.49432}                           & \textbf{0.31252}                              & {\color[HTML]{009901} \textbf{(+)0.01163}}                    \\ \hline
\textbf{Gundapu Sunil}                       & \textbf{0.23573}                       & \textbf{0.23011}                        & \textbf{0.26234}                        & \textbf{0.52132}                           & \textbf{0.31237}                              & {\color[HTML]{009901} \textbf{(+)0.01148}}                    \\ \hline
\textbf{Nowshed CSECU KDE MA}            & \textbf{0.23701}                       & \textbf{0.25460}                        & \textbf{0.25172}                        & \textbf{0.50207}                           & \textbf{0.31135}                              & {\color[HTML]{009901} \textbf{(+)0.01046}}                    \\ \hline
\textbf{Lb732 SESAM}                       & \textbf{0.24276}                       & \textbf{0.24874}                        & \textbf{0.25000}                        & \textbf{0.49910}                           & \textbf{0.31015}                              & {\color[HTML]{009901} \textbf{(+)0.00926}}                    \\ \hline
\textbf{Xiaoyu}                     & \textbf{0.25482}                       & \textbf{0.25415}                        & \textbf{0.24128}                        & \textbf{0.48848}                           & \textbf{0.30969}                              & {\color[HTML]{009901} \textbf{(+)0.00880}}                    \\ \hline
\textbf{NLPU iowa NLP@UIowa}              & \textbf{0.24413}                       & \textbf{0.24348}                        & \textbf{0.23842}                        & \textbf{0.50896}                           & \textbf{0.30875}                              & {\color[HTML]{009901} \textbf{(+)0.00786}}                    \\ \hline
\textbf{Steve050798 NIT-Agartala-NLP-Team} & \textbf{0.25930}                       & \textbf{0.23632}                        & \textbf{0.24507}                        & \textbf{0.48914}                           & \textbf{0.30746}                              & {\color[HTML]{009901} \textbf{(+)0.00657}}                    \\ \hline
\textbf{Sanath97 TDResearch}               & \textbf{0.23006}                       & \textbf{0.23861}                        & \textbf{0.25755}                        & \textbf{0.48907}                           & \textbf{0.30382}                              & {\color[HTML]{009901} \textbf{(+)0.00293}}                    \\ \hline
\textbf{Lyr123 mem3210}                    & \textbf{0.23859}                       & \textbf{0.23448}                        & \textbf{0.24549}                        & \textbf{0.49453}                           & \textbf{0.30327}                              & {\color[HTML]{cc3399} \textbf{(+)0.00238}}                    \\ \hline
\rowcolor[HTML]{9698ED} 
\textbf{Baseline}                          & \textbf{0.24838}                       & \textbf{0.24087}                        & \textbf{0.23019}                        & \textbf{0.48412}                           & \textbf{0.30089}                              & \textbf{}                                                     \\ \hline
{\color[HTML]{cc3399}\textbf{Saradhix Fermi}}                    & \textbf{0.14053}                       & \textbf{0.23262}                        & \textbf{0.26141}                        & {\color[HTML]{cc3399}\textbf{0.53411}}                           & \textbf{0.29217}                              & {\color[HTML]{FE0000} \textbf{(-)0.00872}}                    \\ \hline
\textbf{Hg}                                 & \textbf{0.21494}                       & \textbf{0.19354}                        & \textbf{0.23326}                        & \textbf{0.52218}                           & \textbf{0.29098}                              & {\color[HTML]{FE0000} \textbf{(-)0.00991}}                    \\ \hline
\textbf{LT3}                                & \textbf{0.25142}                       & \textbf{0.17072}                        & \textbf{0.20473}                        & \textbf{0.51410}                           & \textbf{0.28524}                              & {\color[HTML]{FE0000} \textbf{(-)0.01565}}                    \\ \hline
\textbf{Abaruah IIITG-ADBU}                & \textbf{0.22457}                       & \textbf{0.17031}                        & \textbf{0.23535}                        & \textbf{0.50957}                           & \textbf{0.28495}                              & {\color[HTML]{FE0000} \textbf{(-)0.01594}}                    \\ \hline
\textbf{IrinaBejan}                         & \textbf{0.24561}                       & \textbf{0.23506}                        & \textbf{0.25059}                        & \textbf{0.38748}                           & \textbf{0.27969}                              & {\color[HTML]{FE0000} \textbf{(-)0.02120}}                    \\ \hline
\textbf{INGEOTEC-sabino}                    & \textbf{0.23471}                       & \textbf{0.21811}                        & \textbf{0.21950}                        & \textbf{0.43109}                           & \textbf{0.27585}                              & {\color[HTML]{FE0000} \textbf{(-)0.02504}}                    \\ \hline
\textbf{Sabino INGEOTEC}                   & \textbf{0.23471}                       & \textbf{0.21811}                        & \textbf{0.21950}                        & \textbf{0.43109}                           & \textbf{0.27585}                              & {\color[HTML]{FE0000} \textbf{(-)0.02504}}                    \\ \hline
\textbf{Taha IUST}                         & \textbf{0.19378}                       & \textbf{0.19197}                        & \textbf{0.22750}                        & \textbf{0.43149}                           & \textbf{0.26118}                              & {\color[HTML]{FE0000} \textbf{(-)0.03971}}                    \\ \hline
\textbf{Souvik Mishra Kraken}               & \textbf{0.00000}                       & \textbf{0.00000}                        & \textbf{0.00000}                        & \textbf{0.50708}                           & \textbf{0.12677}                              & {\color[HTML]{FE0000} \textbf{(-)0.17412}}                    \\ \hline
\end{tabular}
}
\captionof{Table}{Team/Class--wise results (Macro-F1) and their comparison with the base-line performance, for Task C—Semantic Classification. Table is arranged in descending order ie. top-most row shows the winner while the $2^{nd}$ row tells about the score of $2^{nd}$ ranker. The scores and the team names highlighted in pink color shows the class wise best result.  [Comparison color code: Green--ahead of the base-line; Red--behind the base-line].}
\label{tab:taskc}
\end{table}

\begin{table*}[h]
\centering
\resizebox{1.0\columnwidth}{!}{%
\small{
\begin{tabular}{|c|c|c|c|c|c|c|c|c|c|c|}
\hline
\cellcolor{yellow}\textbf{Team} &\cellcolor{yellow} \textbf{Inception} & \cellcolor{yellow}\textbf{ResNet} &\cellcolor{yellow} \textbf{BERT} & \cellcolor{yellow}\textbf{XLNet} & \cellcolor{yellow}\textbf{LSTM} & \cellcolor{yellow}\textbf{GRU} & \cellcolor{yellow}\textbf{CNN} & \cellcolor{yellow}\textbf{VGG-16} & \cellcolor{yellow}\textbf{DenseNet}& \cellcolor{yellow} \textbf{GloVe} \\ \hline
\textbf{Hitachi} &\cmark&\cmark&\cmark&\cmark&  &  &  &  &  &  \\ \hline
\textbf{YNU-HPCC} &  &\cmark&\cmark&  &\cmark&\cmark&\cmark&  &  &  \\ \hline
\textbf{PRHLT-UPV} &  &  &\cmark&  &\cmark&  &\cmark&\cmark&\cmark&  \\ \hline
\textbf{Guoym} &  &\cmark&\cmark&  &  &\cmark&  &  &  &  \\ \hline
\textbf{Vkeswani IITK} &  &\cmark&\cmark&  &  &  &  &  &  &  \\ \hline
\textbf{Memebusters} &\cmark&  &\cmark&  &\cmark&\cmark&  &  &  &\cmark\\ \hline
\textbf{Sunil Gundapu} &\cmark&  &  &  &  &\cmark&\cmark&  &  &\cmark\\ \hline
\textbf{Suciati Indra} &  &  &  &  &  &  &  &\cmark&  &  \\ \hline
\textbf{SESAM Bonheme} &  &  &  &  &  &  &  &  &  &  \\ \hline
\textbf{Zehao Liu} &  &  &  &  &\cmark&  &\cmark&  &  &  \\ \hline
\textbf{NUAA-QMUL} &  &\cmark&\cmark&  &  &  &  &  &\cmark&  \\ \hline
\textbf{Ambuje Gupta} &  &  &  &  &  &  &  &\cmark&  &  \\ \hline
\textbf{CN-HIT-MI.T} &  &\cmark&\cmark&  &  &  &  &  &  &  \\ \hline
\textbf{KAFK} &  &  &\cmark&  &  &  &  &  &  &  \\ \hline
\textbf{NIT-Agartala-NLP-Team} &  &  &\cmark&  &\cmark&  &  &  &  &\cmark\\ \hline
\textbf{DSC IIT-ISM} &  &\cmark&  &  &\cmark&  &  &  &  &  \\ \hline
\textbf{Sabino Infotech} &\cmark&\cmark&  &  &  &  &  &  &  &  \\ \hline
\textbf{UPB} &  &  &\cmark&  &  &  &  &\cmark&  &  \\ \hline
\textbf{Sravani IS} &  &  &  &  &\cmark&  &\cmark&  &  &  \\ \hline
\textbf{NAYEL} &  &  &  &  &  &  &  &  &  &  \\ \hline
\textbf{IIITG-ADBU} &\cmark&  &\cmark&  &\cmark&  &  &\cmark&  &\cmark\\ \hline
\textbf{LT3} &  &\cmark&  &  &  &  &  &  &  &  \\ \hline
\textbf{Urszula} &  &  &  &  &\cmark&  &\cmark&\cmark&  &\cmark\\ \hline
\textbf{CSECU KDE MA} &  &  &  &  &\cmark&  &  &  &  &  \\ \hline
\textbf{Ingroj Jonathan} &  &  &  &  &\cmark&\cmark&  &  &  & \\ \hline
\textbf{Adithya Sanath} &  &  &\cmark&\cmark&\cmark&  &  &  &  &  \\ \hline
\end{tabular}
}}
\captionof{Table}{Checklist of pre-trained models and techniques, implemented by different teams in their work.}
\label{tab:model_performance}
\end{table*} 

\section{Results, Analysis, and Takeaway points from Memotion 1.0}
\label{sec:result}
Table~\ref{tab:taska}, \ref{tab:taskb} and \ref{tab:taskc} shows the best scores of the all the participants and the comparison with the baseline model whereas Table~\ref{tab:model_performance} shows a summary of the models employed by different participants. Some of the noteworthy points regarding various techniques and consideration of different modals is described in the subsequent sections.

\subsection{Unimodal vs Multi-modal}

\begin{itemize}
    \item \textbf{Considering only text:} Steve and CSECU KDE MA used only textual features to determine the humour as well as the corresponding intensity.  System of steve include a Logistic Regression baseline,  a BiLSTM + Attention-based learner and a transfer learning approach with BERT while CSECU KDE MA applied fastext forward embedding followed by convolution layers with multiple kernel sizes and time distribution LSTM with attention mechanism. Results are quite significant but less than the models implemented considering the combination of image and text.

\item \textbf{Combination of image+text:} Most participant's approach include fusion of visual and textual features extracted using different models as shown in Table \ref{tab:model_performance}. Teams HonoMi Hitachi, Lisa Bonheme and Yingmei Guo proposed ensemble learning where as teams Sunil, DelaPen and many others have used multimodal approaches, few teams even have performed transfer learning on pre-trained models of BERT~\cite{devlin-etal-2019-bert}, VGG-16~\cite{Simonyan2015VeryDC} , ResNet, etc.
\end{itemize}

\subsection{\textbf{Techniques based on}}
\begin{itemize}
\item\textbf{ Visual approaches:} Pretrained models like Inception-ResNet~\cite{DBLP:journals/corr/HeZRS15},Polynet~\cite{zhang2016polynet}, SENet~\cite{DBLP:journals/corr/abs-1709-01507} and the popular off-the-shelf systems like VGG-16~\cite{Simonyan2015VeryDC} and ResNET~\cite{DBLP:journals/corr/HeZRS15} are significantly leveraged as part of the visual feature extractors.

\item \textbf{Textual feature extraction approaches:}
For modeling the content based on textual format, techniques like BiLSTM, BIGRU, and Attention models are used to perform cross domain suggestion mining. Besides these, BiLSTM+Attention based learner and a transfer learning approach with BERT ~\cite{devlin-etal-2019-bert} is also used for analysing the text.
\item \textbf{Approaches to handle categorical imbalance:}
Interestingly, to address the inherent skewness within the categorical data distribution at different levels, approaches like GMM and Training Signal Annealing (TSA) are used.

\end{itemize}

\subsection{Special mentions}
 In addition to the description of top performing models, we have some unique systems implemented by various participants, summarized below:

\begin{itemize}
\item \textbf{Li Zhen hit-mitlab:} Proposed the usage of multiple ways to handle few minor issues of the task such as imbalance and noise. They used RandAugment to enhance the image, and used Training Signal Annealing (TSA) to handle the imbalance. RandAugment detects data augmentation with a reduced search space. They use pre-trained models ResNet-101 and BERT to handle image and text features respectively. They also extracted effective features of the image considering textual information, and concatenate both the obtained features of image and text.
\item \textbf{Bonheme:} Analysed the meme by applying Canonical Correlation Analysis (CCA), Deep Canonical Correlation Analysis (DCCA) that shows no statistically significant correlation between image and
text and observed that image and text are more
complementary than correlated in the case of sentiment analysis of memes. Thus, concluded that alignment based techniques are not suitable for meme analysis. They have used fusion approach with statistical modeling such as random forest and KNN and for the better performance they have used Multi Layer Perceptron(MLP).  With various experiments, they have shown that considering either of the text or image performs better than considering the combination of both.
\item \textbf{Pradyumn Gupta:} Proposed a system which uses different bimodal fusion techniques like GMU, early and late fusion to
leverage the inter-modal dependency for sentiment and emotion classification tasks. To extract visual features, they have used  facial expression, face emotions and different pretrained deep learning models like  ResNet-50~\cite{DBLP:journals/corr/HeZRS15}, AlexNet~\cite{NIPS2012_4824}. To understand the textual information associated with a meme, BiLSTM, RoBERTa are used.
\end{itemize}
\section{Related Work}
\label{sec:Related work}
Identifying the text in the image is as important as the context of the image, so we present the related work in two parts, one involving the models used to extract text from the meme and the other on the analysis of memes.\par
 \cite{jaderberg2014synthetic} proposed one of the first CNN based approach for text recognition to classify words into fixed set of character texts. ~\cite{cacho2019using} uses an n-gram model to correct the OCR text extracted.   ~\cite{memon2020handwritten} performed a comprehensive literature review on handwriting character recognition. A survey analysed in ~\cite{islam2017survey} shows an overview of different aspects of OCR and discuss different methods at resolving issues related to OCR. Template matching using contours is used in ~\cite{olszewska2015active} to recognize visual characters from real-world scenarios.
While there are not many works involving direct classification of emotions on memes. Early works on the detection of offensive content on online social media is OffenseEval ~\cite{ZampieriEA19} shared task, organized at SemEval since 2019. The latest entrant ``Internet memes'' ~\cite{williams2016racial} has doubled the challenge. Detecting an offensive meme is more complex than detecting an offensive text as it involves visual cues and language understanding. Automate of meme generation process are explored in ~\cite{peirson2018dank,oliveira2016one} while others have sought to extract the memes' inherent sentiment ~\cite{french2017image}. Nevertheless, this challenge proves the necessity of more research in the field of multi-modal approaches to detect and classify the sentiments of memes.

\section{Memotion Analysis - the next Horizon!}
\label{sec:conclusion}
The submissions that we received also came along with their respective analytical reasoning, towards ascertaining whether an image or a text or their combination contributes towards modeling the associated emotions from memes. Most of the analyses provided present conflicting views, regarding the importance of a particular content modality. This essentially reinstates the requirement of further investigations into better approaches towards modeling the affect related information from the multi-modal content like memes. The complexity of understanding the emotions from a meme arises primarily due to the interaction of both image and embedded text. Although, few results reported are better when evaluating over either image or text, a human always attempts to take cognizance of both image and text to understand the meaning intended. The challenge is highlighted more by memes which are domain specific, i.e. based on a popular Movie or TV Show. While there are several State-of-the-Art deep learning based systems that leverage data intensive training approaches, that perform various tasks at par with humans on both image and especially for text, still there is a lot more space for applications involving multi-modality like memes, to drive the required progress. Recently Facebook proposed a challenge ~\cite{kiela2020hateful} to classify the meme as Hateful and Not Hateful content.\par
 At present, memes have become one of the most prominent ways of expressing an individual's opinion towards societal issues. Further on classifying the emotion of memes, this work can be extended as follows:
 \begin{itemize}
     \item Properly annotated meme data is still scarce. We plan to enrich our data-set with annotations for different language memes
 (Hinglish, Spanglish etc).  
 \item The success of Memotion motivates us to go further
and organize similar events in the future.
\item The emotion classification could be further extended to develop a meme recommendation system as well as establishing a AI algorithm that could flag the offensive meme from social media platforms automatically.
 \end{itemize}

\bibliographystyle{unsrt}
\bibliography{semeval2020}

\begin{thebibliography}{10}

\bibitem{meme_swift}
Nikhil Sonnad.
\newblock The world’s biggest meme is the word “meme” itself.
\newblock 2018.

\bibitem{gal2016gets}
Noam Gal, Limor Shifman, and Zohar Kampf.
\newblock “it gets better”: Internet memes and the construction of
  collective identity.
\newblock {\em New Media \& Society}, 18(8):1698--1714, 2016.

\bibitem{williams2016racial}
Amanda Williams, Clio Oliver, Katherine Aumer, and Chanel Meyers.
\newblock Racial microaggressions and perceptions of internet memes.
\newblock {\em Computers in Human Behavior}, 63:424--432, 2016.

\bibitem{ZampieriEA19}
Marcos Zampieri, Shervin Malmasi, Preslav Nakov, Sara Rosenthal, Noura Farra,
  and Ritesh Kumar.
\newblock {SemEval}-2019 {T}ask 6: Identifying and categorizing offensive
  language in social media ({OffensEval}).
\newblock In {\em 13th International Workshop on Semantic Evaluation}, pages
  75--86. ACL, 2019.

\bibitem{peirson2018dank}
V~Peirson, L~Abel, and E~Meltem Tolunay.
\newblock Dank learning: Generating memes using deep neural networks.
\newblock {\em arXiv preprint arXiv:1806.04510}, 2018.

\bibitem{oliveira2016one}
Hugo~Gon{\c{c}}alo Oliveira, Diogo Costa, and Alexandre~Miguel Pinto.
\newblock One does not simply produce funny memes!--explorations on the
  automatic generation of internet humor.
\newblock In {\em 7th International Conference on Computational Creativity},
  pages 238--245, Paris, France, 2016. Sony CSL.

\bibitem{french2017image}
Jean~H French.
\newblock Image-based memes as sentiment predictors.
\newblock In {\em 2017 International Conference on Information Society
  (i-Society)}, pages 80--85. IEEE, 2017.

\bibitem{Rosetta}
Viswanath Sivakumar, Albert Gordo, and Manohar Paluri.
\newblock Rosetta: Understanding text in images and videos with machine
  learning.
\newblock 2018.

\bibitem{subjective}
Sicheng Zhao, Guiguang Ding, Tat-Seng Chua, Björn Schuller, and Kurt Keutzer.
\newblock Affective image content analysis: A comprehensive survey.
\newblock pages 5534--5541, 07 2018.

\bibitem{glove}
Jeffrey Pennington, Richard Socher, and Christoper Manning.
\newblock Glove: Global vectors for word representation.
\newblock volume~14, pages 1532--1543, 01 2014.

\bibitem{Simonyan2015VeryDC}
Karen Simonyan and Andrew Zisserman.
\newblock Very deep convolutional networks for large-scale image recognition.
\newblock {\em CoRR}, abs/1409.1556, 2015.

\bibitem{DBLP:journals/corr/HeZRS15}
Kaiming He, Xiangyu Zhang, Shaoqing Ren, and Jian Sun.
\newblock Deep residual learning for image recognition.
\newblock {\em CoRR}, abs/1512.03385, 2015.

\bibitem{NIPS2012_4824}
Alex Krizhevsky, Ilya Sutskever, and Geoffrey~E Hinton.
\newblock Imagenet classification with deep convolutional neural networks.
\newblock In F.~Pereira, C.~J.~C. Burges, L.~Bottou, and K.~Q. Weinberger,
  editors, {\em Advances in Neural Information Processing Systems 25}, pages
  1097--1105. Curran Associates, Inc., 2012.

\bibitem{rahman2019mbert}
Wasifur Rahman, Md~Kamrul Hasan, Amir Zadeh, Louis-Philippe Morency, and
  Mohammed~Ehsan Hoque.
\newblock M-bert: Injecting multimodal information in the bert structure, 2019.

\bibitem{devlin-etal-2019-bert}
Jacob Devlin, Ming-Wei Chang, Kenton Lee, and Kristina Toutanova.
\newblock {BERT}: Pre-training of deep bidirectional transformers for language
  understanding.
\newblock pages 4171--4186, Minneapolis, Minnesota, June 2019. Association for
  Computational Linguistics.

\bibitem{peters2018deep}
Matthew~E. Peters, Mark Neumann, Mohit Iyyer, Matt Gardner, Christopher Clark,
  Kenton Lee, and Luke Zettlemoyer.
\newblock Deep contextualized word representations, 2018.

\bibitem{lan2019albert}
Zhenzhong Lan, Mingda Chen, Sebastian Goodman, Kevin Gimpel, Piyush Sharma, and
  Radu Soricut.
\newblock Albert: A lite bert for self-supervised learning of language
  representations, 2019.

\bibitem{DBLP:journals/corr/SzegedyIV16}
Christian Szegedy, Sergey Ioffe, and Vincent Vanhoucke.
\newblock Inception-v4, inception-resnet and the impact of residual connections
  on learning.
\newblock {\em CoRR}, abs/1602.07261, 2016.

\bibitem{zhang2016polynet}
Xingcheng Zhang, Zhizhong Li, Chen~Change Loy, and Dahua Lin.
\newblock Polynet: A pursuit of structural diversity in very deep networks,
  2016.

\bibitem{DBLP:journals/corr/abs-1709-01507}
Jie Hu, Li~Shen, and Gang Sun.
\newblock Squeeze-and-excitation networks.
\newblock {\em CoRR}, abs/1709.01507, 2017.

\bibitem{DBLP:journals/corr/abs-1712-00559}
Chenxi Liu, Barret Zoph, Jonathon Shlens, Wei Hua, Li{-}Jia Li, Li~Fei{-}Fei,
  Alan~L. Yuille, Jonathan Huang, and Kevin Murphy.
\newblock Progressive neural architecture search.
\newblock {\em CoRR}, abs/1712.00559, 2017.

\bibitem{noauthororeditor}
Alec Radford, Jeffrey Wu, Rewon Child, David Luan, Dario Amodei, and Ilya
  Sutskever.
\newblock Language models are unsupervised multitask learners.
\newblock 2018.

\bibitem{dai2019transformerxl}
Zihang Dai, Zhilin Yang, Yiming Yang, Jaime Carbonell, Quoc~V. Le, and Ruslan
  Salakhutdinov.
\newblock Transformer-xl: Attentive language models beyond a fixed-length
  context, 2019.

\bibitem{DBLP:journals/corr/abs-1906-08237}
Zhilin Yang, Zihang Dai, Yiming Yang, Jaime~G. Carbonell, Ruslan Salakhutdinov,
  and Quoc~V. Le.
\newblock Xlnet: Generalized autoregressive pretraining for language
  understanding.
\newblock {\em CoRR}, abs/1906.08237, 2019.

\bibitem{jaderberg2014synthetic}
Max Jaderberg, Karen Simonyan, Andrea Vedaldi, and Andrew Zisserman.
\newblock Synthetic data and artificial neural networks for natural scene text
  recognition.
\newblock {\em arXiv preprint arXiv:1406.2227}, 2014.

\bibitem{cacho2019using}
Jorge Ram{\'o}n~Fonseca Cacho, Kazem Taghva, and Daniel Alvarez.
\newblock Using the google web 1t 5-gram corpus for ocr error correction.
\newblock In {\em 16th International Conference on Information Technology-New
  Generations (ITNG 2019)}, pages 505--511. Springer, 2019.

\bibitem{memon2020handwritten}
Jamshed Memon, Maira Sami, and Rizwan~Ahmed Khan.
\newblock Handwritten optical character recognition (ocr): A comprehensive
  systematic literature review (slr).
\newblock {\em arXiv preprint arXiv:2001.00139}, 2020.

\bibitem{islam2017survey}
Noman Islam, Zeeshan Islam, and Nazia Noor.
\newblock A survey on optical character recognition system.
\newblock {\em arXiv preprint arXiv:1710.05703}, 2017.

\bibitem{olszewska2015active}
Joanna~Isabelle Olszewska.
\newblock Active contour based optical character recognition for automated
  scene understanding.
\newblock {\em Neurocomputing}, 161:65--71, 2015.

\bibitem{kiela2020hateful}
Douwe Kiela, Hamed Firooz, Aravind Mohan, Vedanuj Goswami, Amanpreet Singh,
  Pratik Ringshia, and Davide Testuggine.
\newblock The hateful memes challenge: Detecting hate speech in multimodal
  memes, 2020.

\end{thebibliography}

\end{document}